\documentclass[journal]{IEEEtran}
\usepackage{amsmath,amsfonts}
\usepackage[linesnumbered,ruled,nosemicolon]{algorithm2e} 
\usepackage{array}
\usepackage[caption=false,font=normalsize,labelfont=sf,textfont=sf]{subfig}
\usepackage{textcomp}
\usepackage{stfloats}
\usepackage{url}
\usepackage{verbatim}
\usepackage{graphicx}
\usepackage{cite}
\usepackage{afterpage}
\usepackage{hyperref}
\usepackage{amsmath}
\usepackage{color}
\usepackage{amssymb}   
\usepackage{cleveref}
\usepackage{booktabs} 
\usepackage{multirow}
\usepackage{siunitx} 
\usepackage{subfig}  
\usepackage{booktabs}    
\usepackage{tabularx}
\usepackage{bm}
\allowdisplaybreaks[4] 

\begin{document}

\title{Efficient Split Federated Learning for Large\\ Language Models over Communication Networks}

\author{Kai Zhao, Zhaohui Yang, Ye Hu, Mingzhe Chen, \IEEEmembership{Senior Member, IEEE}, Chen Zhu, \\Zhaoyang Zhang, \IEEEmembership{Senior Member, IEEE}}



\maketitle

\begin{abstract}
Fine-tuning pre-trained large language models (LLMs) in a distributed manner poses significant challenges on resource-constrained edge networks. To address this challenge, we propose SflLLM, a novel framework that integrates split federated learning with parameter-efficient fine-tuning techniques. By leveraging model splitting and low-rank adaptation (LoRA), SflLLM reduces the computational burden on edge devices. Furthermore, the introduction of a federated server facilitates parallel training and enhances data privacy. To accommodate heterogeneous communication conditions and diverse computational capabilities of edge devices, as well as the impact of LoRA rank selection on model convergence and training cost, we formulate a joint optimization problem of both communication and computation resource. The formulated problem jointly optimizes subchannel allocation, power control, model splitting point selection, and LoRA rank configuration, aimed at minimizing total training delay. An iterative optimization algorithm is proposed to solve this problem efficiently. Specifically, a greedy heuristic is employed for subchannel allocation, the power control subproblem is reformulated as a convex optimization problem using auxiliary variables, and an exhaustive search is adopted for optimal split position and rank selection. Simulation results demonstrate that the proposed SflLLM framework achieves comparable model accuracy while significantly reducing client-side computational requirements. Furthermore, the proposed resource allocation scheme and adaptive LoRA rank selection strategy notably reduce the training latency compared to conventional approaches.
\end{abstract}

\section{Introduction}

\IEEEPARstart{T}{he} rapid advancement of pre-trained models has has significantly advanced fields such as computer vision and natural language processing \cite{achiam2023gpt,chowdhery2023palm}. According to scaling laws \cite{kaplan2020scaling}, increasing the number of model parameters and training data volume typically leads to improved model performance. Meanwhile, with the proliferation of lnternet of things (IoT) devices and sensors within wireless networks, edge devices have become capable of generating and collecting vast amounts of data. Effectively leveraging this distributed data can further enhance model performance. However, due to privacy concerns and communication constraints \cite{thirunavukarasu2023large,wu2023bloomberggpt}, transmitting raw data to a central server is often impractical. To overcome these challenges, federated learning (FL) has emerged as a promising paradigm for the distributed fine-tuning of large language models (LLMs) \cite{fan2023fate,kuang2024federatedscope,ye2024openfedLLM}. FL preserves privacy and reduces communication overhead by transmitting model parameters rather than raw data \cite{konevcny2016federated}.

Nevertheless, as the size of models continues to increase, fully loading and fine-tuning these large-scale models on resource-limited edge devices remains computationally infeasible. To mitigate this challenge, split learning (SL) has been proposed \cite{vepakomma2018split}, which partitions the model and offloads computationally intensive tasks to a more resourceful server. Instead of exchanging entire model parameters, SL allows clients to transmit only intermediate activations and corresponding gradients, thereby substantially reducing communication overhead \cite{lin2024efficient,lin2024split}. However, a critical drawback of SL is its dependence on sequential interactions between clients and the server, severely restricting the overall training efficiency. To address this limitation, split federated learning (SFL) has been recently introduced, effectively combining the privacy preservation and parallelization advantages of FL with the communication efficiency of SL \cite{thapa2022splitfed}. Consequently, SFL is emerging as a highly promising framework for distributed training across heterogeneous edge computing environments.

Fine-tuning pre-trained models on downstream tasks typically yields substantial improvements in performance. However, as the scale of model parameters increases, performing full fine-tuning becomes computationally prohibitive due to intensive resource requirements. While model quantization provides a promising approach for reducing computational overhead by compressing models, it typically requires specialized hardware capable of low-bit computations. A more practical and hardware-compatible approach is low-rank adaptation (LoRA) \cite{hu2022lora,sheng2023s}, which introduces trainable low-rank modules into the existing model structure, significantly decreasing the number of trainable parameters while maintaining comparable performance to full fine-tuning. When combined with SFL, LoRA further reduces client-side computational workload and the volume of uploaded data, enabling more communication-efficient fine-tuning of large-scale models. Although prior studies has explored the integration of FL with LoRA, the combination of LoRA specifically with SFL remains under-explored in existing literature\cite{lin2024splitlora}.

Although resource-rich cloud environments can easily accommodate the computational demands associated with large-scale models, resource-constrained edge devices typically lack such computational capacities. Therefore, minimizing computational overhead becomes essential for effective deployment in these edge computing scenarios. Furthermore, the straggler effect prevalent in SFL can significantly hinder training progress, thus efficient resource allocation strategies are vital for reducing latency and maintaining training efficiency. Additionally, existing studies often focus solely on a single training round and neglecting the comprehensive impact of model training hyperparameters on the total latency across the entire training process. Moreover, SFL involves extensive parameter exchanges between clients and two separate servers, where computation and communication are inherently coupled. Thus, jointly optimizing computation and communication processes represents a necessary and promising perspective for enhancing the efficiency of the training process\cite{10915662}.

In this paper, we propose SflLLM, a novel split federated learning framework specifically designed for efficiently fine-tuning LLMs within resource-constrained edge networks. SflLLM integrates LoRA and SFL techniques to achieve efficient distributed training. The proposed framework leverages distributed co-training mechanisms to securely and effectively harness data dispersed across edge devices, while the use of model splitting significantly alleviates local resource requirements. Moreover, the incorporation of efficient fine-tuning techniques reduces both computational burden and communication overhead.

Nevertheless, due to the inherent heterogeneity in channel conditions and computational capabilities among edge devices, the emergence of stragglers can substantially degrade overall training efficiency \cite{chen2021distributed,lee2017speeding}. A appropriately designed resource allocation and model splitting strategy can mitigate the adverse impacts of stragglers and accelerate training speed, thereby enhancing overall learning efficiency. Additionally, the rank chosen for the LoRA module influences both computation complexity and communication overhead, thereby affecting the convergence speed of the training process. Proper selection of LoRA rank strikes a balance between these overheads and can significantly reduce total training delay.

To address these challenges, we formulate a comprehensive optimization problem for SflLLM that involves subchannel assignment, power control, split layer selection, and rank selection. Subsequently, we develop an efficient algorithm to solve the formulated optimization problem. The primary contributions of this paper are summarized as follows:

\begin{itemize}
\item We propose the SflLLM framework for fine-tuning LLMs in resource-constrained edge networks. Specifically, the model is partitioned between the main server and clients, significantly reducing computational load and memory usage at the edge side. A federated server is introduced to aggregate client models while effectively preserving privacy. By employing LoRA techniques, the scale of trainable parameters is minimized, substantially alleviating the computational burden on edge devices and reducing communication overhead for uploads to the federated server, thus realizing a distributed framework for fine-tuning LLMs with efficient communication and computation.

\item We theoretically analyze how the rank of the LoRA module affects training overhead and convergence speed within the SflLLM framework. To achieve efficient model training, we design a joint optimization strategy for resource allocation and task offloading, specifically targeting subchannel allocation, power control, model split location, and LoRA rank selection. The objective of this strategy is to minimize the training latency while meeting the total power budget requirements.

\item To address this problem, we propose an iterative algorithm that alternately solves subproblems related to subchannel assignment, power control, split location, and rank selection. Specifically, the subchannel assignment subproblem is solved using a greedy algorithm, the power allocation subproblem via convex optimization techniques, and both the split location and rank selection subproblems are addressed through exhaustive search methods.

\item Extensive simulation demonstrate that the proposed SflLLM framework effectively accomplishes fine-tuning tasks with high efficiency. Furthermore, the customized resource allocation schemes and rank selection strategies substantially reduce training latency compared to traditional methods.
\end{itemize}

The remainder of this paper is organized as follows: Section \ref{Related Work} reviews related work. Sections \ref{System Model} and \ref{Framework} introduce the system model and framework of SflLLM, respectively. Sections \ref{Resource Allocation} and \ref{Solution} detail the formulation of the resource management problem and the corresponding solution algorithm. Simulation results and analysis are provided in Section \ref{Evaluation}. Finally, the conclusions are drawn in Section \ref{Conclusion}.

\section{Related Work}\label{Related Work}
FL has emerged as a widely adopted distributed learning framework, which effectively preserves data privacy while maintaining communication efficiency. However, effectively deploying FL in edge network, where client devices have limited computational and communication resources, remains a critical challenge for improving performance \cite{shi2022toward}. Extensive research has addressed this limitations, focusing on model compression techniques\cite{chen2023service}, optimized resource allocation \cite{jiao2020toward}, hierarchical model aggregation \cite{chen2022federated}, and effective client selection strategies \cite{xu2020client}. SL, another distributed learning paradigm, mitigates some FL limitations\cite{konevcny2016federated} by offloading the majority of the training workload to the server side\cite{lin2024split}. This significantly reduces computational cost and communication overhead at the client. However, early SL implementations \cite{vepakomma2018split} required clients to sequentially perform training, resulting in low training efficiency. Split Federated Learning (SFL) has been introduced to overcome this limitation \cite{thapa2022splitfed}. By combining the benefits of both SL and FL, SFL has become a promising framework for distributed learning. Motivated by its advantages, recent research has extensively explored various aspects of SFL, including model splitting strategies \cite{wu2023split}, model aggregation mechanisms \cite{lin2024adaptsfl}, and client selection methods tailored for wireless environments \cite{liu2022wireless}.

The effectiveness of SFL on small-scale models has been demonstrated in several recent studies. For instance, a hierarchical SFL framework was proposed in \cite{khan2023joint} to address challenges such as single points of failure, fairness issues, and slow convergence rates, achieving notable performance on the MNIST dataset. For client heterogeneity, a ring-topology-based SFL was developed in \cite{shen2023ringsfl}, which demonstrated better convergence performance than benchmark methods on both the independent and identically distributed (IID) and non-IID datasets. This approach has been successfully applied to mainstream networks, including  ResNet18 \cite{he2016deep}, VGG16 \cite{simonyan2014very}, and AlexNet \cite{krizhevsky2012imagenet}, and also effectively prevents eavesdroppers from recovering training data. However, SFL for LLMs remains an area that has not been fully explored. Some research has begun to focus on fine-tuning LLMs using FL architectures for distributed collaborative training over wireless networks\cite{fan2023fate}. \cite{jiang2024low} introduces a low-parameter federated learning framework employing cost-effective parameter construction, local model fine-tuning, and global model aggregation using soft labeling and LoRA modules. Additionally, \cite{fan2024fedcoLLM} employs lightweight adapters within LLMs to facilitate knowledge exchange between servers and clients, thereby maintaining data privacy while minimizing overheads. Although these studies emphasize lightweight techniques for efficient training and reduced client workload, they fail to adequately address the resource constraints on clients through model splitting techniques. In this paper, we provide an efficient resource-constrained fine-tuning approach for LLMs by combining SFL with LoRA.

SFL involves extensive parameter exchanges over wireless networks, making it both a significant and challenging problem to effectively utilize the resources available on edge devices and central servers. In \cite{xu2023accelerating}, the authors addressed a real-world scenario with heterogeneous devices and a single split point in a deep neural network. They formulated and solved a joint optimization problem of split-point selection and bandwidth allocation to minimize system latency. Additionally, \cite{zhu2024esfl} proposed an efficient parallel split learning algorithm that jointly optimizes client-side workloads and server-side resource allocation while considering user heterogeneity. In \cite{khan2023joint}, the hierarchical SFL framework was introduced, solving an optimization problem that accounted for device-relative local accuracy, wireless resource allocation, task offloading, and transmit power control, aiming to minimize latency and energy consumption.Although these studies provide effective solutions for resource allocation in traditional deep neural network-based SFL, they do not specifically address the unique requirements of fine-tuning LLMs or adapting pre-trained models. To address this gap, in this paper, we propose an optimization framework that jointly considers communication resource allocation, task offloading, and fine-tuning hyperparameters for LLMs, explicitly accounting for the impact of LoRA on SFL overheads.

\begin{figure}[!t]
\centering
\includegraphics[width=\linewidth]{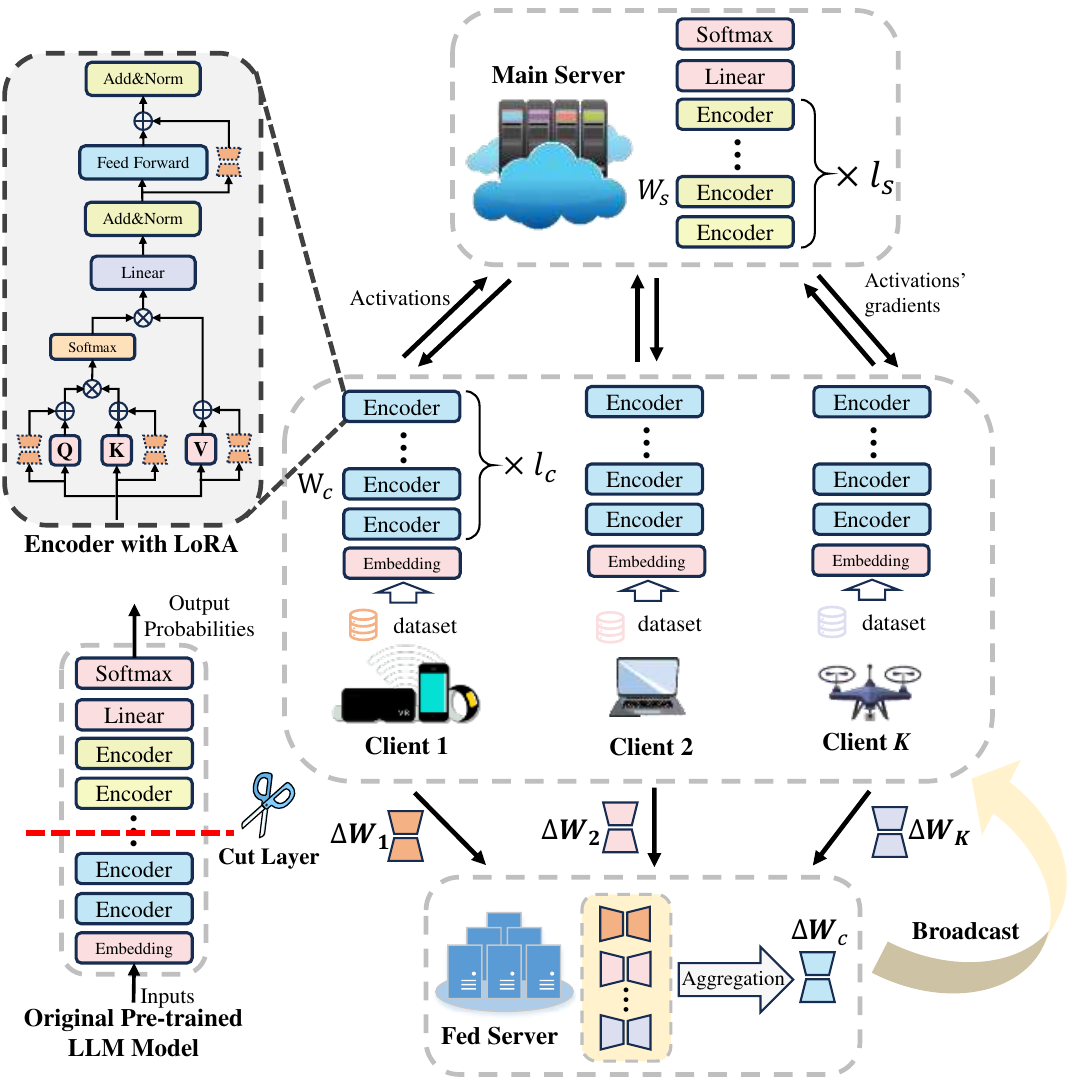}
\caption{Illustration of the proposed SflLLM framework over wireless networks.}
\label{system_model}
\end{figure}

\section{System Model}\label{System Model}
In this section, we describe the system model of the proposed SflLLM framework, which lays the groundwork for subsequent analysis and discussion. As depicted in Fig.~\ref{system_model}, a typical scenario for deploying SflLLM involves the following three main components:

\begin{itemize} 
\item\textbf{Client:} Clients typically consist of edge devices with limited computational capabilities, such as medical IoT devices. The first portion of the model is deployed on these device, enabling each client to perform local forward and backward propagation. Let $\mathcal{K} = \{1, 2, \dots, K\}$ represent the set of participating clients, where $K$ is the total number of clients. The client $k$ holds a local dataset $\mathcal{D}_{k}=\{\bm{x}_{k,i},y_{k,i}\}_{i=1}^{D_{k}}$, where $D_{k}$ denotes the number of data samples. $\bm{x}_{k,i}$ and $y_{k,i}$ represent the $i$-th input and its corresponding label, respectively. We denote the pre-trained model parameters deployed on the client by $\bm{W}_{c}$, and the trainable parameters introduced through the LoRA module as $\Delta \bm{W}_{c}$.

\item \textbf{Main Server:} The main server is equipped with abundant computational and communication resources. It is responsible for executing the forward and backward propagation for the latter portion of the model. The pre-trained model parameters residing at the main server are denoted as $\bm{W}_{s}$, and the trainable parameters introduced by the LoRA module are represented by $\Delta \bm{W}_{s}$.

\item\textbf{Federated Server:} The federated server aggregates the client-side trainable parameters after $I$ rounds of local training steps, subsequently updating them to form a new global client model. For privacy enhancement, the federated server and main server are typically maintained by separate entities. This physical and logical separation prevents attackers from easily reconstructing original client data, even if parameters from both servers are compromised simultaneously.
\end{itemize}

To evaluate the model performance across different learning tasks, we define the loss function $f(\bm{W}_{c/s}, \Delta \bm{W}_{c/s}, \bm{x}_{k,i}, y_{k,i})$, which measures the discrepancy between the model predictions and the ground-truth labels for a given input vector $\bm{x}_{k,i}$ with corresponding label $y_{k,i}$, under the current model parameters. For different learning tasks, the loss function will be different. For example, $f(\bm{W}_{c/s}, \Delta \bm{W}_{c/s}, \bm{x}_{k,i}, y_{k,i}) = \frac{1}{2}(\bm{x}_{k,i}^T(\bm{W}_{c/s}+\Delta \bm{W}_{c/s}))-y_{k,i})^2$ for linear regression and $f(\bm{W}_{c/s}, \Delta \bm{W}_{c/s}, \bm{x}_{k,i}, y_{k,i}) = -\log(1+\exp(-y_{k,i}\bm{x}_{k,i}^T(\bm{W}_{c/s}+\Delta \bm{W}_{c/s})))$ for logistic regression. The loss function for the $k$-th client can be expressed as:
\begin{align}
F_{k}(\bm{W}_{c}, \Delta \bm{W}_{c}, \bm{W}_{s}, \Delta \bm{W}_{s}) 
&= \frac{1}{D_{k}} \sum_{i=1}^{D_{k}} 
     f(\bm{W}_{c}, \Delta \bm{W}_{c}, \nonumber\\
&\quad\; \bm{W}_{s}, \Delta \bm{W}_{s}, 
     \bm{x}_{k,i}, y_{k,i}) 
\end{align}
where $D_k$ denotes the number of local data samples held by client $k$.

The global learning objective of the proposed SflLLM framework is to minimize the total loss aggregated across all participating clients, which can be formulated as:

\begin{align}
\min_{\Delta \bm{W}_{c}, \Delta \bm{W}_{s}} 
& F(\bm{W}_{c}, \Delta \bm{W}_{c}, \bm{W}_{s}, \Delta \bm{W}_{s}) \notag \\
&\hspace{-2em}= \frac{1}{D} \sum_{k=1}^{K} \sum_{i=1}^{D_k} 
    f(\bm{W}_{c}, \Delta \bm{W}_{c}, \bm{W}_{s}, \Delta \bm{W}_{s}, 
    \bm{x}_{k,i}, y_{k,i}) \label{eq2}
\end{align}
where $D = \sum_{k=1}^{K} D_k$ denotes the total number of training samples across all clients. For clarity and ease of reference, the key notations used throughout this paper are summarized in Table~\ref{tab:notations}.

\begin{table}[!t]
  \caption{Summary of Key Notations}
  \label{tab:notations}
  \centering
  \small  
  \renewcommand{\arraystretch}{1.15}  
  \begin{tabularx}{\linewidth}{@{}lX@{}}
    \toprule
    \textbf{Notation} & \textbf{Description} \\
    \midrule
    $K,\;\mathcal{K}$ & The number and set of participating clients  \\
    $\mathcal{D}_k, D_k$ & Local dataset of client $k$ and its number  \\
    $E(r)$ & Steps to reach target loss (depends on rank $r$)  \\
    $I$ & Local steps per global round  \\
    $b$ & Batch size  \\
    $\ell_c/\ell_s$ & The number of layers deployed on client/main server\\
    $\bm{W}_c/\bm{W}_s$ & Frozen pre‑trained weights on client/main server  \\
    $\Delta\bm{W}_c/\Delta\bm{W}_s$ & LoRA trainable weights on client/main server  \\
    $r$ & LoRA rank  \\
    $f_k/f_s$ &The computing capability of client $k$/main server  \\
    $\kappa_k/\kappa_s$ & GPU cycles required per FLOP on client/main server  \\
    $\rho_j/\varpi_j$ &The FP/BP computation workload of pre‑trained weights at layer $j$ per sample \\
    $\Delta\rho_j/\Delta\varpi_j$ &The FP/BP computation workload of LoRA weights at layer $j$ per rank for one sample \\
    $\psi_j$ & The data size of activations at layer $j$  \\
    $\Phi_c^{F/B}, \Delta\Phi_c^{F/B}$ & Client‑side FP/BP computation workload per sample for pre‑trained and trainable weights\\
    $\Phi_s^{F/B}, \Delta\Phi_s^{F/B}$ & Server‑side FP/BP computation workload per sample for pre‑trained and trainable weights\\
    $\Gamma_s$ & The data size of activations per sample\\
    $\Delta\Theta_c$ & The data size of client‑side LoRA weights \\
    $M/N, \mathcal{M}/\mathcal{N}$ & The number and set of subchannels to main/federated server  \\
    $r_k^{i}, p_i, \mu_j$ & Decision variables (explained in Section \ref{Resource Allocation})  \\
    $B_i^{s}, B_i^{f}$ & The bandwidth of $i$-th subchannel   \\
    $G_c/G_s/G_f$ & The effective antenna gain of client/main server/federated server  \\
    $\gamma(d_k^{s/f})$ & The average channel gain from client $k$ to main/federated server  \\
    $(\sigma^{s/f})^{2}$ & The PSD of the noise from client $k$ to main/federated server\\
    $p_k^{\max}, p_{th}^{s/f}$ & The maximum transmit PSD of client $k$ and main/federated server \\
    $\theta_{k,\xi}^{s}, \theta_{k,\xi}^{f}$ & Auxiliary rate variables after log‑convexification  \\
    $T_1, T_2, T_3$ & Auxiliary variables: $\max_k(T_k^{F}+T_k^{s})$, $\max_k T_k^{B}$, $\max_k T_k^{f}$  \\
    \bottomrule
  \end{tabularx}
\end{table}

\section{The SflLLM Framework}\label{Framework}

This section presents a detailed description of the workflow for the proposed SflLLM framework. Before initiating the model training, the central server initializes the model by partitioning it into two distinct segments: the client-side segment and the server-side segment. Each global training round consists of two main phases: first, each client and the main server conduct $I$ rounds of local fine-tuning, followed by the federated server aggregating local models from clients to update the global model. This iterative process continues until convergence is achieved.

\subsection{Fine-Tuning Phase}

The fine-tuning process conducted at each client and the main server during a single training round can be detailed through the following steps:

\paragraph{Client-Side Forward Propagation} In this step, each participating client $k$ randomly selects a mini-batch $\mathcal{B}_k \subseteq \mathcal{D}_k$ from its local dataset, where $\mathcal{B}_k$ contains $B_k$ data samples. Each client independently performs forward propagation (FP) on its assigned portion of the model. After completing the client-side forward propagation, activation vectors are generated at the split layer (i.e., the final layer of the client-side model). The activation vector produced by the $k$-th client at the $t$-th training step can be expressed as: 
\begin{equation}\label{eq3} 
\bm{s}_k^t=\varphi\left(\bm{W}_c, \Delta\bm{W}_{c,k}^{t-1}, \bm{x}_k^t\right), 
\end{equation} 
where $\varphi(\bm{W},\Delta\bm{W},\bm{x})$ denotes the mapping function between input data $\bm{x}$ and the predicted output, given the pre-trained parameters $\bm{W}$ and LoRA trainable parameters $\Delta\bm{W}$.

\paragraph{Uploading of Activation Vectors} After completing the client-side forward propagation, all participating clients upload their activation vectors $\bm{s}_k^t$ and the corresponding labels $y_k^t$ to the main server via wireless communication channels. Subsequently, the main server leverages these collected activation vectors to further fine-tuning the server-side model.

\paragraph{Server-Side Forward Propagation} Upon receiving the activation vectors, the main server inputs these vectors into the server-side model to perform forward propagation. Let $\Delta\bm{W}_{s}^{t-1}$ denote the LoRA trainable parameters of the server-side model at the $(t-1)$-th training step. Thus, the predicted output of the main server at the $t$-th step can be expressed as: 
\begin{equation}\label{eq4} 
\hat{y}_k^t=\varphi\left(\bm{W}_{s},\Delta\bm{W}_{s}^{t-1},\bm{S}^t\right), 
\end{equation} 
where $\bm{S}^{t}=[\bm{s}_{1}^{t};\bm{s}_{2}^{t};\dots;\bm{s}_{K}^{t}]$ represents the aggregated activation vectors from all participating clients. After forward propagation, the main server calculates the loss by comparing the predicted outputs to their corresponding true labels across all received samples.

\paragraph{Server-Side Backward Propagation} Based on the computed loss, the main server executes backward propagation (BP), starting from the output layer, to compute gradients for the trainable parameters within the server-side LoRA module. These parameters are updated according to: \begin{equation}\label{eq5} 
\Delta\bm{W}_s^{t}\leftarrow\Delta\bm{W}_s^{t-1}-\eta_s\bm{G}_{s}^{t}, 
\end{equation} 
where $\bm{G}_{s}^{t}$ represents the gradient of trainable parameters at the $t$-th training step, and $\eta_s$ denotes the learning rate of the server-side model.

\paragraph{Downloading of Activation Vector Gradients} After completing the BP process, the main server transmits the gradients of the activation vectors with respect to the client-side trainable parameters, denoted by $\frac{\partial \bm{s}_{k}^{t}}{\partial \Delta \bm{W}_{c,k}^{t}}$, back to the corresponding client over the wireless network.

\paragraph{Client-Side Backward Propagation} Each client subsequently performs backward propagation to fine-tune its local trainable parameters based on the received gradients of the activation vectors. The parameter update rule for client $k$ at the $t$-th training step is described by: 
\begin{equation}\label{eq6} 
\Delta \bm{W}_k^{t} \leftarrow \Delta \bm{W}_k^{t-1} - \eta_c \bm{G}_k^t, 
\end{equation} 
where $\bm{G}_k^t$ represents the gradients of the trainable parameters at the $t$-th training step on client $k$, and $\eta_c$ denotes the learning rate for the client-side model.

\subsection{Aggregation Phase} 
Once the clients and main server have completed $I$ rounds of fine-tuning, the federated server initiates the aggregation phase to aggregate and update the local models collected from all participating clients. This phase consists of the following three sequential steps:

\paragraph{Uploading Client-Side LoRA Modules} Each participating client uploads the trainable parameters of its client-side LoRA module to the federated server through wireless communication.

\paragraph{Aggregation of Client-Side LoRA Modules} Upon receiving the LoRA modules from all clients, the federated server aggregates these modules to construct a new global client-side model as described below:
\begin{equation}\label{eq7} 
\Delta \bm{W}_c^{t} = \sum_{k=1}^K \frac{{D}_k}{{D}} \Delta \bm{W}_k^{t}. 
\end{equation} 

\paragraph{Broadcasting the Global Client-Side LoRA Module} After completing the aggregation step, the federated server broadcasts the newly aggregated global client-side LoRA module to all participating clients over the wireless network. Each client then updates its local LoRA module accordingly, preparing for the subsequent fine-tuning round.

The complete training procedure of the proposed SflLLM framework is summarized in Algorithm~\ref{alg:cap}.

\begin{algorithm}[t]
\small
\caption{The SflLLM Training Framework.}\label{alg:cap}
\KwIn{%
$E$, $I$, $\gamma_c$, $\gamma_s$, $\bm{x}_k^t$, ${y}_k^t$, $\mathcal{K}$, $\bm{W}_c$, $\bm{W}_s$, $\mathcal{D}_k$, $\mathcal{D}$}
\KwOut{$\Delta\bm{W}_{c}^{*}$, $\Delta\bm{W}_{s}^{*}$}
\BlankLine
Initialize $\Delta\bm{W}_{c}^{0}$,$\Delta\bm{W}_{s}^{0}$\;
\For{$t = 1, 2, \ldots, E$}{
    \tcp{Runs on clients}
    \ForEach{$ k \in \mathcal{K}$ \textbf{in parallel}}{
        $\bm{s}_k^t=\varphi\left(\bm{W}_c,\Delta\bm{W}_{c,k}^{t-1},\bm{x}_k^t\right)$\;
        Send $(\bm{s}_k^t,{y}_k^t)$ to main server\;
    }\;
    
    \tcp{Runs on main server}
    $\bm{S}^{t}=[\bm{s}_{1}^{t};\bm{s}_{2}^{t};...;\bm{s}_{K}^{t}]$\;
    ${\hat{y}}_k^t=\varphi\left(\bm{W}_{s},\Delta\bm{W}_{s}^{t-1},\bm{S}^t\right)$\;
    Calculate loss function value $F(\bm{W}_{c},\Delta\bm{W}_{c},\bm{W}_{s},\Delta\bm{W}_{s})$\;
    Calculate gradients of server-side model $\bm{G}_{s}^{t}$\;
    $\Delta\bm{W}_s^{t}\leftarrow\Delta\bm{W}_s^{t-1}-\gamma_s\bm{G}_{s}^{t}$\;
    Broadcast aggregated activations' gradients $\frac{\partial \bm{s}_{k}^{t}}{\partial\Delta\bm{W}_{c,k}^{t}}$ to all clients\;
    \;
    
    \tcp{Runs on clients}
    \ForEach{$ k \in \mathcal{K}$ \textbf{in parallel}}{
        Calculate gradients of client-side model $\bm{G}_{k}^{t}$\;
        $\Delta\bm{W}_k^{t}\leftarrow\Delta\bm{W}_k^{t-1}-\gamma_c\bm{G}_{k}^{t}$\;
    }\;

    \tcp{Runs on fed server}
    \If{$t \bmod I = 0$}{
        $\Delta\bm{W}_c^{t}=\sum_{k=1}^K\frac{|\mathcal{D}_k|}{|\mathcal{D}|}\Delta\bm{W}_{k}^{t}$\; 
        Broadcast the new global client-side LoRA adapter $\Delta\bm{W}_c^{t}$ to all clients\;
    }
}
\end{algorithm}

\section{Resource Allocation for SflLLM}\label{Resource Allocation}

This section formulates the resource allocation problem within the proposed SflLLM framework, with the objective of minimizing the total training delay. Due to heterogeneous local computing capabilities and dynamically varying communication resources among different clients, significant discrepancies can arise in individual training speeds. These variations often result in stragglers, substantially increasing the overall training latency. Therefore, efficient resource allocation is critical to mitigating delays by balancing workload and communication resources among clients. Furthermore, the choice of model split point directly impacts both the amount of activation vector data exchanged and the computational load at each client. Thus, selecting an optimal split point adapted to the deployment environment is essential for minimizing training delay.

With the integration of the LoRA module, both the computational load and communication overhead increase proportionally to the LoRA rank $r$. Within a certain range, adjusting the rank significantly affects the convergence rate. Specifically, increasing the LoRA rank generally reduces the number of training steps required to reach the target accuracy, thereby influencing the overall training latency. Consequently, determining an optimal rank is vital for efficiently accelerating model training and minimizing total latency.

To address these challenges comprehensively, we propose an efficient resource allocation strategy that jointly optimizes split layer selection, LoRA rank selection, subchannel assignment, and transmit power control. By considering these factors collectively, the proposed method effectively minimizes the training delay within the SflLLM framework.

\subsection{Training Delay Model}

To rigorously formulate the resource allocation problem, we first define the key resource variables involved as follows:

\begin{itemize} 
\item[$\bullet$] $\bm{r} \colon$Let binary variables $r_{k}^{i,s} \in \{0,1\}$ and $r_{k}^{i,f} \in \{0,1\}$ represent the subchannel allocation decisions. Specifically, $r_{k}^{i,s}=1$ indicates that the $i$-th subchannel between the client and the main server is allocated to client $k$; otherwise, $r_{k}^{i,s}=0$. Similarly, $r_{k}^{i,f}=1$ indicates that the $i$-th subchannel between the client and the federated server is allocated to client $k$; otherwise, $r_{k}^{i,f}=0$. The corresponding allocation vectors for client $k$ are denoted as $\bm{r}_k^s = [r_k^{1,s}, r_k^{2,s}, \ldots, r_k^{M,s}]$ and $\bm{r}_k^f = [r_k^{1,f}, r_k^{2,f}, \ldots, r_k^{N,f}]$, where $M$ and $N$ denote the total numbers of subchannels available for communications between the clients and the main server, and between the clients and the federated server, respectively. Hence, $\bm{r}^s = [\bm{r}_1^s, \bm{r}_2^s, \ldots, \bm{r}_K^s]$ and $\bm{r}^f = [\bm{r}_1^f, \bm{r}_2^f, \ldots, \bm{r}_K^f]$ represent the complete subchannel allocation matrices.

\item[$\bullet$] $\bm{p} \colon$Let $p_i^s \geq 0$ denote the transmission power spectral density (PSD) of the $i$-th subchannel used for uploading activations or gradients from clients to the main server. Similarly, $p_i^f \geq 0$ represents the transmission PSD of the $i$-th subchannel used by clients when uploading local model parameters to the federated server. $\bm{p}^s = [p_1^s, p_2^s, \ldots, p_M^s]$ and $\bm{p}^f = [p_1^f, p_2^f, \ldots, p_N^f]$ represent transmission power allocation decisions.

\item[$\bullet$] $\boldsymbol{\mu} \colon$Let binary variables $\mu_j \in \{0,1\}$ represent the decision of the split point between client and main server for each layer $j$. Specifically, $\mu_j=1$ indicates that the $j$-th layer is deployed at the client side, and $\mu_j=0$ indicates deployment at the main server side. Thus, the split configuration is expressed as a binary vector $\boldsymbol{\mu} = [\mu_1, \mu_2, \ldots, \mu_{\ell_c + \ell_s}]$, where $\ell_c$ and $\ell_s$ denote the number of layers available at the client and server, respectively.

\item[$\bullet$] $r \colon$Let the integer variable $r \geq 1$ denote the rank of the LoRA module.
\end{itemize}

Split federated learning proceeds through multiple rounds of training until the model reaches convergence. Each complete forward and backward propagation process constitutes a local iteration round, which is sequentially executed by all participating clients and the main server. After $I$ local iteration rounds, clients upload their local model parameters to the federated server for aggregation. Simultaneously, the main server aggregates the local models from different clients to generate a new global model. This process completes one global iteration round, and convergence is achieved after $E$ global rounds.

For simplicity, we focus on analyzing a single local training round, omitting the index $t$ for training rounds in the subsequent discussion. The latency for transmitting activation vector gradients from the main server to the client, as well as for broadcasting the global model by the federated server, is negligible due to the generally high transmission power of servers and the relatively small size of the transmitted data. Additionally, the aggregation latency on both the federated server and the main server is also minimal and can be disregarded, as the computational requirements for these operations are relatively low, and the servers have ample resources. Consequently, we focus on the latency of the remaining six phases within a training round.

Assuming a pre-trained weight matrix $W \in \mathbb{R}^{d \times k}$, LoRA approximates the weight update as the product of two matrices through low-rank decomposition: $W_0 + \Delta W = W_0 + BA$, where $A \in \mathbb{R}^{r \times k}$, $B \in \mathbb{R}^{d \times r}$, and the rank $r \ll \min(d, k)$. During training, $W_0$ remains frozen (i.e., no updates to the weights), while $A$ and $B$ are the trainable weights. Therefore, the number of parameters introduced by the LoRA module is $r \times (d + k)$, which scales linearly with $r$.

\subsubsection{Clients Perform Forward Propagation}
In this phase, all clients perform FP in parallel. Let $\Phi_c^F(\boldsymbol{\mu}) = \sum_{j=1}^{\ell_c + \ell_s} \mu_j \rho_j$ represent the computational workload (in FLOPs) of the client's pre-trained model for processing each sample during the FP phase. Here, $\rho_j$ indicates the computational workload for the $j$-th layer of the pre-trained model. With the introduction of the LoRA module, the number of trainable parameters increases, resulting in an increased computational workload. Let $\Delta \Phi_c^F(\boldsymbol{\mu}, r) = \sum_{j=1}^{\ell_c + \ell_s} \mu_j r \Delta \rho_j$ denote the computational workload corresponding to the trainable parameters introduced by the LoRA module, where $\Delta \rho_j$ is the computational workload for the forward propagation of the trainable parameters at the $j$-th layer for each data sample per rank.

Each client randomly selects a mini-batch of $b$ samples for processing during one round of forward propagation. The FP delay for client $k$ can be expressed as:
\begin{equation}\label{eq8} 
T_k^F = \frac{b \kappa_k \left( \Phi_c^F(\boldsymbol{\mu}) + \Delta \Phi_c^F(\boldsymbol{\mu}, r) \right)}{f_k}, \forall k \in \mathcal{K}, 
\end{equation}
where $f_k$ denotes the computational capability of client $k$ (i.e., the number of graphics processing unit (GPU) cycles per second), and $\kappa_k$ is the number of GPU cycles required to complete a single floating-point operation on client $k$.

\subsubsection{Transmission of Activation Vectors}
After completing the forward propagation, each client sends the activation vector from the last layer to the main server over the wireless channel. We consider a frequency division multiple access (FDMA) communication scenario. The activation vectors output by the LoRA module and the pre-training parameters are combined. Therefore, the size of the activation vector data remains unchanged despite the introduction of the LoRA module. Let $\Gamma_s(\boldsymbol{\mu}) = \sum_{j=1}^{\ell_c + \ell_s - 1} (\mu_j - \mu_{j+1}) \psi_j$ denote the data size of the activation vectors in bits, where $\psi_j$ is the data size for the activation vectors at the $j$-th layer of the pre-trained model. The uplink transmission rate from client $k$ to the main server can be expressed as:

\begin{equation}\label{eq9} 
R_k^s = \sum_{i=1}^{M} r_k^{i,s} B_i^s \log_2 \left( 1 + \frac{p_i^s G_c G_s \gamma(d_k^s)}{(\sigma^s)^2} \right), \forall k \in \mathcal{K}, 
\end{equation}
where $B_i^s$ is the bandwidth allocated to the $i$-th subchannel between the client and the main server, $G_c$ and $G_s$ respectively represent the effective antenna gains of the client and main server, $\gamma(d_k^s)$ is the average channel gain from client $k$ to the main server, $d_k^s$ is the communication distance between client $k$ and the main server, and $(\sigma^s)^2$ is the PSD of the wireless channel noise. The activation vector transmission delay for client $k$ can be expressed as:
\begin{equation}\label{eq10} 
T_k^s = \frac{b \Gamma_s(\boldsymbol{\mu})}{R_k^s}, \forall k \in \mathcal{K}. 
\end{equation}

\subsubsection{Main Server Performs Forward Propagation}

In this phase, the main server receives activation vectors from all participating clients and performs forward propagation. Let $\Phi_s^F(\boldsymbol{\mu}) = \sum_{j=1}^{\ell_c + \ell_s} (1 - \mu_j) \rho_j$ represent the computational workload of the main server's pre-trained model for processing activation vectors from a single client, and $\Delta \Phi_s^F(\boldsymbol{\mu}, r) = \sum_{j=1}^{\ell_c + \ell_s} (1 - \mu_j) r \Delta \rho_j$ represent the computational workload corresponding to the trainable parameters introduced by the LoRA module. The main server uses the activation vectors uploaded by all participating clients to perform forward propagation. The forward propagation delay on the main server can be expressed as:
\begin{equation}\label{eq11} 
T_s^F = \frac{K b \kappa_s \left( \Phi_s^F (\boldsymbol{\mu}) + \Delta \Phi_s^F (\boldsymbol{\mu}, r) \right)}{f_s}, 
\end{equation}
where $f_s$ denotes the computational capability of the main server (i.e., the number of GPU cycles per second), and $\kappa_s$ denotes the number of GPU cycles required by the main server to complete a single floating-point operation.

\subsubsection{Main Server Performs Backward Propagation}
After completing forward propagation, the main server initiates backward propagation. Let $\Phi_s^B(\boldsymbol{\mu}) = \sum_{j=1}^{\ell_c + \ell_s} (1 - \mu_j) \varpi_j$ represent the total computational workload for backward propagation of the pre-trained parameters for each data sample, where $\varpi_j$ is the computational workload for the $j$-th layer of the neural network. Let $\Delta \Phi_s^B(\boldsymbol{\mu}, r) = \sum_{j=1}^{\ell_c + \ell_s} (1 - \mu_j) r \Delta \varpi_j$ represent the total computational workload for backward propagation of the trainable parameters in the main server, where $\Delta \varpi_j$ is the computational workload associated with the trainable parameters at the $j$-th layer for each data sample and rank. The backward propagation delay on the main server can be expressed as:
\begin{equation}\label{eq12} 
T_s^B = \frac{K b \kappa_s \left( \Phi_s^B (\boldsymbol{\mu}) + \Delta \Phi_s^B (\boldsymbol{\mu}, r) \right)}{f_s}. 
\end{equation}

\subsubsection{Client Performs Backward Propagation}
In this phase, each client receives the gradient of the activation vector and performs the backward propagation process. Let $\Phi_c^B(\boldsymbol{\mu}) = \sum_{j=1}^{\ell_c + \ell_s} \mu_j \varpi_j$ denote the total computational workload for the client’s pre-trained parameters during backward propagation for each data sample. Let $\Delta \Phi_c^B(\boldsymbol{\mu}, r) = \sum_{j=1}^{\ell_c + \ell_s} \mu_j r \Delta \varpi_j$ represent the computational workload for the trainable parameters. The backward propagation delay for client $k$ can be expressed as:
\begin{equation}\label{eq13} 
T_k^B = \frac{b \kappa_k \left( \Phi_c^B (\boldsymbol{\mu}) + \Delta \Phi_c^B (\boldsymbol{\mu}, r) \right)}{f_k}, \forall k \in \mathcal{K}.
\end{equation}

\subsubsection{Client Local Model Upload}
After completing $I$ rounds of local training, each client uploads its trainable model parameters to the federated server for aggregation into the global model. While certain non-trainable parameters may be aggregated locally, all trainable parameters from the client-side model are transmitted to the federated server. The updated global model is then broadcast to all clients to initiate the next round of training. Since the number of trainable parameters increases linearly with the rank $r$, we define the total data volume of the client-side trainable parameters as $\Delta \Theta_c(\boldsymbol{\mu}, r) = \sum_{j=1}^{\ell_c + \ell_s} \mu_j r \Delta \xi_j$, where $\Delta \xi_j$ denotes the data volume corresponding to the trainable parameters at the $j$-th layer.

The uplink transmission rate from client $k$ to the federated server is expressed as:
\begin{equation}\label{eq14} 
R_k^f = \sum_{i=1}^{N} r_k^{i,f} B_i^f \log_2 \left( 1 + \frac{p_i^f G_c G_f \gamma(d_k^f)}{(\sigma^f)^2} \right), \forall k \in \mathcal{K}, 
\end{equation}
where $B_i^f$ denotes the bandwidth allocated to the $i$-th subchannel, $G_f$ represents the effective antenna gain at the federated server, $\gamma(d_k^f)$ is the average channel gain over the communication distance $d_k^f$, and $(\sigma^f)^2$ denotes the PSD of the channel noise.

The transmission delay for client $k$ to upload its model parameters to the federated server is given by:
\begin{equation}\label{eq15} 
T_k^f = \frac{\Delta \Theta_c (\boldsymbol{\mu}, r)}{R_k^f}, \forall k \in \mathcal{K}. 
\end{equation}

\begin{figure}[t]
\centering
\includegraphics[width=\linewidth]{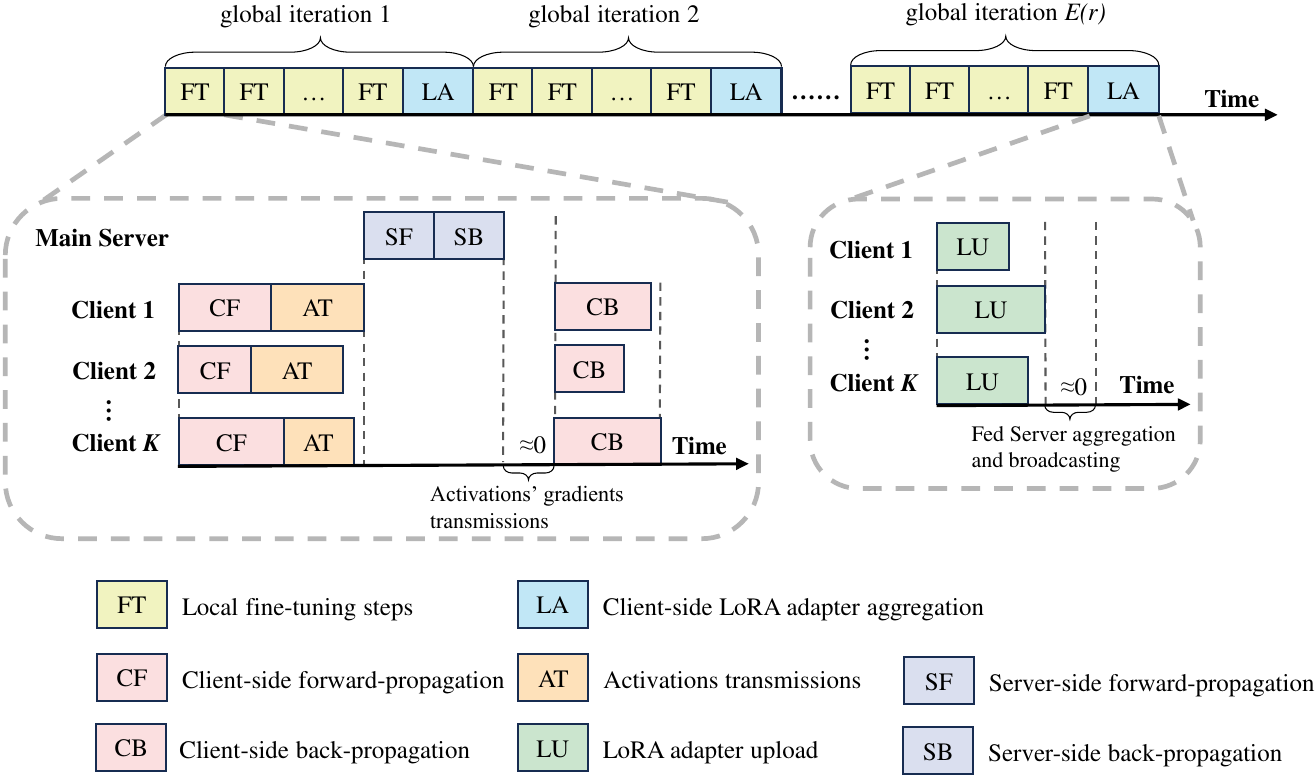}
\caption{Illustration of SflLLM training procedure.}
\label{workflow}
\end{figure}

\subsection{Resource Allocation Problem Formulation}

As depicted in Fig.~\ref{workflow}, the total latency for one round of local training is defined as:
\begin{align}\label{eq16}
T_{local}\bigl(\bm{r}^s,\bm{r}^f,\bm{p}^s,\bm{p}^f,\boldsymbol{\mu},r\bigr)
= &\max_k\{T_k^F + T_k^s\} \;+\; T_s^F \;+\; T_s^B\notag\\
& + \max_k\{T_k^B\}.
\end{align}

The total training delay across all global rounds is given by:
\begin{equation}\label{eq17}
T\left(\bm{r}^s,\bm{r}^f,\bm{p}^s,\bm{p}^f,\boldsymbol{\mu},r\right)=E(r)(IT_{local}+\max_k\{T_k^f\}),
\end{equation}
where $E(r)$ as a function of the rank $r$ denotes the number of global iterations required to achieve the desired accuracy.

Time-varying and heterogeneous wireless channel conditions, coupled with disparities in client computing capabilities, can lead to significant dropout events. Furthermore, the selection of the model split location has a pronounced impact on both computational and communication latency. Consequently, efficient communication resource allocation is critical for accelerating the training process. Based on these insights, we formulate the following optimization problem to minimize the overall training delay:
\begin{subequations}\label{problem}
\begin{alignat}{1}
\mathcal{P}:&\min_{\bm{r}^s,\bm{r}^f,\bm{p}^s,\bm{p}^f,\boldsymbol{\mu},r}T \tag{\ref{problem}}\\
&\begin{aligned}\text{s.t.C1:}&r_k^{i_x,x}\in\{0,1\},\forall k\in\mathcal{K},i_s\in\mathcal{M},i_f\in\mathcal{N},x\in\{s,f\},\end{aligned} \nonumber \\
&\begin{aligned}\text{C2}:\sum_{k=1}^Kr_k^{i_x,x}=1,\quad\forall i_s\in\mathcal{M},i_f\in\mathcal{N},x\in\{s,f\},\end{aligned} \nonumber\\
&\text{C3} \begin{aligned}:&\mu_{j}\in\left\{0,1\right\},\mu_{j}\geq\mu_{j+1},\quad \forall j\in\mathcal{L},\end{aligned} \nonumber\\
&\text{C4} \begin{aligned}:\sum_{i=1}^Mr_k^{i,s}p_i^sB_i^s\leq p_k^{\max},\sum_{i=1}^Nr_k^{i,f}p_i^fB_i^f\leq p_k^{\max},\forall k\in\mathcal{K},\end{aligned} \nonumber\\
&\text{C5} \begin{aligned}:&\sum_{i=1}^M\sum_{k=1}^Kr_k^{i,s}p_i^sB_i^s\leq p_{th}^s,\sum_{i=1}^N\sum_{k=1}^Kr_k^{i,f}p_i^fB_i^f\leq p_{th}^f,\end{aligned} \nonumber\\
&\text{C6} :p_{i_x}^x\geq0,\quad\forall i_s\in\mathcal{M},i_f\in\mathcal{N},x\in\{s,f\},\nonumber\\
&\text{C7} :r\in\mathbb{Z}^+.   \nonumber
\end{alignat}
\end{subequations}
where $\mathcal{M}=\{1,\dots,M\}$ and $\mathcal{N}=\{1,\dots,N\}$ denote the sets of subchannels allocated to the main server and the federated server, respectively. In addition, $\mathcal{L}=\{1,\dots,\ell_c+\ell_s\}$ represents the set of all layers in the neural network.

Constraints C1 and C2 ensure that each subchannel is exclusively assigned to a single edge device, thereby avoiding co-channel interference. Constraint C3 guarantees the uniqueness of the selected model split location. Constraint C4 imposes an upper bound on the transmit power of each client, while C5 enforces the total uplink transmit power constraints at the main and federated servers, characterized by thresholds $p_{th}^s$ and $p_{th}^f$, respectively. Constraint C6 ensures that the transmit power for each subchannel remains non-negative, and C7 enforces that the rank is a positive integer.

\section{Algorithm Design}\label{Solution}
\subsection{Subchannel Assignment}
To address the optimization problem in \eqref{problem}, we begin by fixing the other decision variables and focus on the subchannel assignment subproblem. The resulting subproblem is formulated as:
\begin{subequations}\label{p1}
\begin{alignat}{1}
\mathcal{P}1: & \min_{\bm{r}^s,\bm{r}^f} T \tag{\ref{p1}}\\
& \mathrm{s.t.C1,C2,C4,C5.}  \nonumber
\end{alignat}
\end{subequations}

Problem \eqref{p1} involves a non-convex objective function and integer-valued decision variables, posing significant challenges for direct optimization. As the objective is determined by the maximum latency incurred by client dropouts during uplink transmission to either the main or federated server, minimizing training latency necessitates prioritizing subchannel allocation to clients experiencing higher delays. 

To this end, we propose a greedy subchannel allocation algorithm. The key idea is to assign subchannels with higher transmission rates to clients with weaker computational or communication capabilities at the outset. In each iteration, once every client has been assigned at least one subchannel, the remaining unallocated subchannels are iteratively distributed to the lagging clients—those with the highest observed latency—until all channels are allocated. The complete subchannel allocation procedure is detailed in Algorithm~\ref{alg:greedy_subchannel_allocation}.

\begin{algorithm}[t]
\small
\caption{Greedy Subchannel Allocation Approach.}
\label{alg:greedy_subchannel_allocation}
\KwIn{$\mathcal{K},B_i^s,B_i^f,f_k,d_k^f$}
\KwOut{$\bm{r}^{s*},\bm{r}^{f*}$}
\textbf{Initialization:} Set $\bm{r}^{s}, \bm{r}^{f} \gets 0$\\
The set of remaining clients $\mathcal{A}_s,\mathcal{A}_f \gets \mathcal{K}$\\
The set of subchannels to be allocated $\mathcal{M}\leftarrow\{1,2,\ldots,M\},\mathcal{N}\leftarrow\{1,2,\ldots,N\}$.\\
\textbf{Phase 1: Ensure Each Client Receives at Least One Subchannel} \\
\For{$j = 1, 2, \dots, K$}{
    Find $n\leftarrow\arg\min_{k\in\mathcal{A}_s}\{f_k\}$,$m\leftarrow\arg\max_{i\in\mathcal{M}}\{{B_{i}^s}\}$ \\
    Let $r_n^{m,s}\leftarrow1,\mathcal{A}_s\leftarrow\mathcal{A}_s-\{n\},\mathcal{M}\leftarrow\mathcal{M}-\{m\}$ \\
    Find $n\leftarrow\arg\max_{k\in\mathcal{A}_f}\{d_k^f\}$,$m\leftarrow\arg\max_{i\in\mathcal{N}}\{{B_{i}^f}\}$ \\
    Let $r_n^{m,f}\leftarrow1,\mathcal{A}_f\leftarrow\mathcal{A}_f-\{n\},\mathcal{N}\leftarrow\mathcal{N}-\{m\}$ \\
}

\textbf{Phase 2: Allocate Remaining Subchannels} \\
$\mathcal{A}_s,\mathcal{A}_f \gets \mathcal{K}$ \\
\While{$\mathcal{M}\neq\varnothing$}{
    Find $n\leftarrow\operatorname{arg}\operatorname*{max}_{i\in\mathcal{A}_s}\{T_k^F+T_k^s\}$, $m\leftarrow\arg\max_{i\in\mathcal{M}}\{{B_{i}^s}\}$ \\
    \If{$\bm{r}_n$ does not meet C4 or C5 in \eqref{problem} }{
        $\mathcal{A}_s\leftarrow\mathcal{A}_s-\{n\}$ \\
    }\Else{
        $r_n^m\leftarrow1$,Update $T_k^s,\mathcal{M}\leftarrow\mathcal{M}-\{m\}$ \\
    }
}
\While{$\mathcal{N}\neq\varnothing$}{
    Find $n\leftarrow\operatorname{arg}\operatorname*{max}_{i\in\mathcal{A}_f}\{T_k^f\}$, $m\leftarrow\arg\max_{i\in\mathcal{N}}\{{B_{i}^f}\}$ \\
    \If{$\bm{r}_n$ does not meet C4 or C5 in \eqref{problem} }{
        $\mathcal{A}_f\leftarrow\mathcal{A}_f-\{n\}$ \\
    }\Else{
        $r_n^m\leftarrow1$,Update $T_k^f,\mathcal{N}\leftarrow\mathcal{N}-\{m\}$ \\
    }
}
\end{algorithm}

\subsection{Power Control}
After completing the subchannel assignment for each client, we introduce auxiliary variables $p_{k,\xi}^s$ and $p_{k,\xi}^f$ to represent the transmission PSD of the $\xi$-th subchannel allocated to client $k$ for uplink transmissions to the main server and the federated server, respectively. Likewise, $B_{k,\xi}^s$ and $B_{k,\xi}^f$ denote the bandwidth of the $\xi$-th subchannel assigned to client $k$ for communication with the main and federated servers, respectively. Given that the subchannel assignment variables $\bm{r}^s$ and $\bm{r}^f$ are fixed, the original problem in \eqref{problem} can be reformulated as:

\begin{subequations}\label{problem2}
\begin{alignat}{1}
& \min_{\bm{p}^s,\bm{p}^f,\boldsymbol{\mu},r}T \tag{\ref{problem2}}\\
& \mathrm{s.t.C3} \nonumber\\
& \widetilde{\mathrm{C}}4:\sum_{\xi=1}^{M_k}p_{k,\xi}^s B_{k,\xi}^s\leq p_k^{\max},\sum_{\xi=1}^{N_k}p_{k,\xi}^f B_{k,\xi}^f\leq p_k^{\max},\quad\forall k\in\mathcal{K}, \nonumber\\
& \widetilde{\mathrm{C}}5:\sum_{k=1}^K\sum_{\xi=1}^{M_k}p_{k,\xi}^s B_{k,\xi}^s\leq p_{th}^s,\sum_{k=1}^K\sum_{\xi=1}^{N_k}p_{k,\xi}^f B_{k,\xi}^f\leq p_{th}^f,\nonumber\\
& \widetilde{\mathrm{C}}6:p_{k,\xi_x}^x\geq0,\quad\forall k\in\mathcal{K}, \xi_s\in\mathcal{M}_k,\xi_f\in\mathcal{N}_k,x\in\{s,f\},  \nonumber\\
&\text{C7} :r\in\mathbb{Z}^+,\nonumber
\end{alignat}
\end{subequations}
where $\mathcal{M}_k = \{1, 2, \ldots, M_k\}$ and $\mathcal{N}_k = \{1, 2, \ldots, N_k\}$ denote the sets of subchannels allocated to client $k$ for communication with the main and federated servers, respectively, and $M_k$, $N_k$ represent the corresponding numbers of allocated subchannels.

The objective function in \eqref{problem2} is non-convex due to the presence of the maximum delay terms, which introduces challenges in deriving an optimal solution. To facilitate tractable optimization, we introduce auxiliary variables $T_1$, $T_2$ and $T_3$, which satisfy $T_1 \geq \max_k \{ T_k^F + T_k^s \}$, $T_2 \geq \max_k \{ T_k^B \}$, and $T_3 \geq \max_k \{ T_k^f \}$. With these substitutions, the problem in \eqref{problem2} can be equivalently transformed into the following form:
\begin{subequations}\label{problem3}
\begin{alignat}{1}
&\min_{\bm{p}^s,\bm{p}^f,\boldsymbol{\mu},r,T_{1},T_{2},T_{3}}\widetilde{T} \tag{\ref{problem3}} \\
& \mathrm{s.t.C3,\widetilde{C}4,\widetilde{C}5,\widetilde{C}6,C7} \nonumber\\
& \mathrm{C8}:\frac{b \kappa_k(\Phi_c^F(\boldsymbol{\mu})+\Delta\Phi_c^F(\boldsymbol{\mu},r))}{f_k}\nonumber\\
&+\frac{b\Gamma_s(\boldsymbol{\mu})}{\sum_{\xi=1}^{M_k}B_{k,\xi}^s\mathrm{log}_{2}\left(1+\frac{p_{k,\xi}^{s}G_{c}G_{s}\gamma(d_{k}^{s})}{(\sigma^{s})^{2}}\right)}\leq T_1,\quad \forall k\in\mathcal{K}, \nonumber\\
& \mathrm{C9}:\frac{b \kappa_k(\Phi_c^B (\boldsymbol{\mu})+\Delta\Phi_c^B (\boldsymbol{\mu},r))}{f_k}\leq T_2,\quad \forall k\in\mathcal{K}, \nonumber\\
& \mathrm{C10}:\frac{\Delta\Theta_{c}(\boldsymbol{\mu},r)}{\sum_{\xi=1}^{N_k}B_{k,\xi}^{f}\mathrm{log}_{2}\left(1+\frac{p_{k,\xi}^{f}G_{c}G_{f}\gamma(d_{k}^{f})}{(\sigma^{f})^{2}}\right)}\leq T_3,\quad \forall k\in\mathcal{K}\nonumber,
\end{alignat}    
\end{subequations}
where $\widetilde{T} (\bm{p}^s, \bm{p}^f, \boldsymbol{\mu}, r, T_1, T_2, T_3) = E(r) ( I (T_1 + T_s^F + T_s^B + T_2) + T_3)$.

To minimize $\widetilde{T}$, it is necessary to determine the optimal values of $T_1$, $T_2$, and $T_3$. The optimal solution of problem~\eqref{problem3} must satisfy $T_1 = \max_k \{ T_k^F + T_k^s \}$, $T_2 = \max_k \{ T_k^B \}$, and $T_3 = \max_k \{ T_k^f \}$ to ensure that the reformulated problem in~\eqref{problem3} remains equivalent to the original formulation in~\eqref{problem2}. However, the non-convex nature of constraints C8 and C10 makes direct optimization intractable.
To address this issue, we introduce auxiliary variables $\boldsymbol{\theta}^s = \{\theta_{1,1}^s, \theta_{1,2}^s, \ldots, \theta_{K,M_K}^s\}$ and $\boldsymbol{\theta}^f = \{\theta_{1,1}^f, \theta_{1,2}^f, \ldots, \theta_{K,N_K}^f\}$, defined as:
\begin{align}\label{theta}
\theta_{k,\xi}^x = B_{k,\xi}^x \log_2( 1 + \frac{p_{k,\xi}^x G_c G_s \gamma(d_k^x)}{(\sigma^x)^2} ),\quad\forall x\in\{s,f\}.
\end{align}

By substituting the auxiliary variables into problem~\eqref{problem3}, the reformulated problem can be expressed as:
\begin{subequations}\label{problem4}
\begin{alignat}{1}
&\min_{\boldsymbol{\theta}^s,\boldsymbol{\theta}^f,\boldsymbol{\mu},r,T_{1},T_{2},T_{3}}\widetilde{T} \tag{\ref{problem4}} \\
& \mathrm{s.t.C3,C7,C9} \nonumber\\
& \hat{\mathrm{C}}4:\sum_{\xi=1}^{M_{k}}\sigma_s^{2}B_{k,\xi}^s\frac{2^{\frac{\theta_{k,\xi}^s}{B_{k,\xi}^s}}-1}{G_{c}G_{s}\gamma(d_{k}^s)}\leq p_{k}^{\mathrm{max}},\sum_{\xi=1}^{N_{k}}\sigma_f^{2}B_{k,\xi}^f\frac{2^{\frac{\theta_{k,\xi}^f}{B_{k,\xi}^f}}-1}{G_{c}G_{f}\gamma(d_{k}^f)}\nonumber \\
&\leq p_{k}^{\mathrm{ma x}},\quad\forall k\in\mathcal{K}, \nonumber \\
& \hat{\mathrm{C}}5:\sum_{k=1}^{K}\sum_{\xi=1}^{M_{k}}\sigma_s^{2}B_{k,\xi}^s\frac{2^{\frac{\theta_{k,\xi}^s}{B_{k,\xi}^s}}-1}{G_{c}G_{s}\gamma(d_{k}^s)}\leq p_{th}^{s},\nonumber \\
&\sum_{k=1}^{K}\sum_{\xi=1}^{N_{k}}\sigma_f^{2}B_{k,\xi}^f\frac{2^{\frac{\theta_{k,\xi}^f}{B_{k,\xi}^f}}-1}{G_{c}G_{f}\gamma(d_{k}^f)}\leq p_{th}^{f}, \nonumber \\
&\hat{\mathrm{C}}6:\theta_{k,\xi_x}^x\geq0,\quad\forall k\in\mathcal{K},\xi_s\in\mathcal{M}_k,\xi_f\in\mathcal{N}_k\nonumber,x\in\{s,f\}, \\
& \hat{\mathrm{C}}8:\frac{b \kappa_k(\Phi_c^F(\boldsymbol{\mu})+\Delta\Phi_c^F(\boldsymbol{\mu},r))}{f_k}+\frac{b\Gamma_s(\boldsymbol{\mu})}{\sum_{\xi=1}^{M_k}\theta_{k,\xi}^s}\leq T_1, \forall k\in\mathcal{K}, \nonumber\\
& \hat{\mathrm{C}}10:\frac{\Delta\Theta_{c}(\boldsymbol{\mu},r)}{\sum_{\xi=1}^{N_k}\theta_{k,\xi}^f}\leq T_3,\quad\forall k\in\mathcal{K}\nonumber,
\end{alignat}    
\end{subequations}

We observe that if the decision variables $\boldsymbol{\mu}$ and $r$ are fixed, the optimization problem~\eqref{problem4} can be simplified to:
\begin{subequations}\label{p2} 
\begin{alignat}{1} 
\mathcal{P}2: & \min_{\boldsymbol{\theta}^s,\boldsymbol{\theta}^f,T_{1},T_{2},T_{3}}\widetilde{T} \tag{\ref{p2}} \\ & \text{s.t.}\quad\hat{\text{C}}4, \hat{\text{C}}5, \hat{\text{C}}6, \hat{\text{C}}8, \text{C}9, \hat{\text{C}}10. \nonumber \end{alignat}
\end{subequations}

The objective function $\widetilde{T}$ in problem \eqref{p2} is a linear combination of $T_{1}$, $T_{2}$, and $T_{3}$, and thus is convex. Constraints $\hat{\mathrm{C}}4$ and $\hat{\mathrm{C}}5$ are convex constraints since their left-hand sides represent linear combinations of exponential functions with respect to variables $\boldsymbol{\theta}^s$ and $\boldsymbol{\theta}^f$. Constraints $\hat{\mathrm{C}}6$ and $\mathrm{C}9$ are obviously convex. Constraints $\hat{\mathrm{C}}8$ and $\hat{\mathrm{C}}10$ are convex as well, since their left-hand sides involve inverse functions of linear combinations of variables $\boldsymbol{\theta}^s$ and $\boldsymbol{\theta}^f$, while their right-hand sides are affine functions of $T_1$ and $T_3$. Therefore, the optimization problem in \eqref{p2} is convex and can be efficiently solved by standard convex optimization solvers such as CVX.

\subsection{Joint Splitting Point Selection and Rank Configuration}
Similarly, if $\boldsymbol{\theta}^s$, $\boldsymbol{\theta}^f$, $r$, $T_1$, $T_2$, and $T_3$ are fixed, problem~\eqref{problem4} reduces to:
\begin{subequations}\label{p3} 
\begin{alignat}{1} 
\mathcal{P}3: & \min_{\boldsymbol{\mu}}\widetilde{T} \tag{\ref{p3}} \\ & \text{s.t.}\quad\mathrm{C}3,\hat{\mathrm{C}}8,\mathrm{C}9,\hat{\mathrm{C}}10. \nonumber 
\end{alignat}
\end{subequations}

Since the number of candidate split locations determined by $\boldsymbol{\mu}$ is generally equal to the number of layers in the network, which is typically small, problem~\eqref{p3} can be solved via exhaustive search.

Furthermore, by fixing $\boldsymbol{\theta}^s$, $\boldsymbol{\theta}^f$, $\boldsymbol{\mu}$, $T_1$, $T_2$, and $T_3$, the original problem~\eqref{problem4} can be further simplified to:
\begin{subequations}\label{p4} \begin{alignat}{1} \mathcal{P}4: & \min_{r}\widetilde{T} \tag{\ref{p4}} \\ & \text{s.t.}\quad\text{C}7, \hat{\text{C}}8, \text{C}9, \hat{\text{C}}10, \nonumber \end{alignat}
\end{subequations}
where the number of global training rounds $E(r)$ can be estimated offline through pretraining on a representative dataset. Since the rank $r$ is typically constrained to a small set of integers, exhaustive search is a practical and effective approach.

Based on the above decomposition, the original optimization problem~\eqref{problem} is divided into four subproblems (i.e., $\mathcal{P}1$, $\mathcal{P}2$, $\mathcal{P}3$, and $\mathcal{P}4$) each solvable independently. Leveraging this structure, we propose a block coordinate descent (BCD)-based algorithm utilizing an alternating optimization strategy to solve the overall problem. The detailed procedure is outlined in Algorithm~\ref{alg:bcd}. Although the non-convex, mixed-integer nature of problem~\eqref{problem} precludes formal convergence guarantees, extensive empirical evaluations demonstrate that the proposed BCD algorithm reliably converges to a stable and effective solution within a finite number of iterations, regardless of initialization.

\subsection{Complexity Analysis}
The complexity of solving problem \eqref{problem} arises from iteratively addressing four subproblems at each iteration. For the subchannel assignment subproblem, the complexity is given by $\mathcal{O}(K(M+N))$. Regarding the power allocation subproblem, its complexity is primarily determined by solving the convex optimization problem \eqref{p2}. Specifically, the complexity of problem \eqref{p2} is $\mathcal{O}(X_1^2 X_2)$~\cite{lobo1998applications}, where $X_1=M+N+3$ denotes the number of optimization variables, and $X_2=5K+2$ is the number of constraints. Hence, the complexity of solving the power control subproblem is $\mathcal{O}(K(M+N)^2)$. For the splitting point selection subproblem, the complexity is $\mathcal{O}(\ell_c+\ell_s)$. For the rank selection subproblem, the complexity is $\mathcal{O}(R)$, where $R$ represents the predefined number of candidate ranks for exhaustive search. Consequently, the total complexity of the proposed resource allocation algorithm is given by $\mathcal{O}\left(\tau_{\text{max}}\left(K(M+N)^2 + (\ell_c+\ell_s) + R\right)\right)$,where $\tau_{\text{max}}$ denotes the number of outer iterations.Since $M$ is typically much larger than $N$, and $KM^2$ is significantly greater than $(\ell_c+\ell_s)$ and $R$, the complexity of the proposed resource allocation algorithm can be simplified to $\mathcal{O}\left(\tau_{\text{max}}K M^2 \right)$.

\begin{algorithm}[t]
\caption{BCD-Based Algorithm for Minimizing Training Delay.}
\label{alg:bcd}
\KwIn{convergence tolerance $\epsilon$, maximum iterations $\tau_{\text{max}}$}
\KwOut{$\bm{r}^{s*}$,$\bm{r}^{f*}$,$\bm{p}^{s*}$, $\bm{p}^{f*}$, $\boldsymbol{\mu}^{*}$, $r^{*}$}
\textbf{Initialization:} $\boldsymbol{\theta}^{s,0}$, $\boldsymbol{\theta}^{f,0}$, $\boldsymbol{\mu}^0$, $r^0$, $T_1^0$, $T_2^0$, $T_3^0$   \\
set iteration index $\tau = 0$  \\
\Repeat{$\left| \widetilde{T}^\tau - \widetilde{T}^{\tau-1} \right| \leq \epsilon$ or $\tau = \tau_{\text{max}}$}{
    $\tau \gets \tau + 1$  \\
    Solve $\mathcal{P}1$ based on Algorithm \ref{alg:greedy_subchannel_allocation} to obtain $\bm{r}^{s,\tau}$, $\bm{r}^{f,\tau}$  \\
    Solve $\mathcal{P}2$ with convex optimization tools to obtain $\boldsymbol{\theta}^{s,\tau}, \boldsymbol{\theta}^{f,\tau},T_1^\tau$, $T_2^\tau$, $T_3^\tau$  \\
    Solve $\mathcal{P}3$ using exhaustive search to update $\boldsymbol{\mu}^\tau$ \\ 
    Solve $\mathcal{P}4$ with exhaustive search to update $r^\tau$  \\
}
\Return{$\bm{r}^{s*}$,$\bm{r}^{f*}$,$\bm{p}^{s*}$, $\bm{p}^{f*}$, $\boldsymbol{\mu}^{*}$, $r^{*}$}
\end{algorithm}

\section{Performance Evaluation}\label{Evaluation}
This section presents simulation results to evaluate the learning performance of the proposed SflLLM framework, alongside the effectiveness of the proposed rank selection and resource allocation strategies.

\subsection{Simulation Setup}
We simulate a system comprising $K$ clients uniformly distributed within a circular area of radius $d_{\max} = \SI{20}{\meter}$, with a federated server positioned at the center and a main server located \SI{100}{\meter} from the centroid. Unless otherwise stated, the number of clients is set to $K=5$. Each client has a computational capability randomly selected from $[1,1.6]$ GHz, while the main server operates at a fixed capacity of \SI{5}{GHz}.

The total bandwidth allocated for client communication with both the federated and main servers is \SI{500}{kHz}, equally divided among subchannels. The noise power spectral density is \SI{-174}{dBm/Hz}~\cite{hu2019edge}, and the maximum transmit power per client is \SI{41.76}{dBm}. Clients share a computing intensity of $1/1024$ cycles/FLOP, whereas the main server is set at $1/32768$ cycles/FLOP.

The path loss is modeled by $128.1 + 37.6\log_{10}(d)$, where $d$ is in kilometers, with a shadow fading standard deviation of \SI{8}{dB}. Simulation parameters are summarized in Table~\ref{tab:sim_params}.

\begin{table}[!ht]
\caption{Simulation Parameters}
\label{tab:sim_params}
\begin{tabular}{@{}llll@{}}
\toprule
\textbf{Parameter} & \textbf{Value} & \textbf{Parameter} & \textbf{Value} \\
\midrule
$f_s$            & $5\ \mathrm{GHz}$               & $f_k$          & $[1.0, 1.6]\ \mathrm{GHz}$ \\
$K$              & $5$                             & $M,N$          & $20$ \\
$\eta_c,\eta_s$  & $4 \times 10^{-4}$              & $G_cG_s$       & $160$ \\
$G_cG_f$         & $80$                            & $\sigma^2_s,\sigma^2_f$  & $-174\ \mathrm{dBm/Hz}$ \\
$\kappa_s$       & $\frac{1}{32768}\ \mathrm{cycles/FLOP}$ & $\kappa_k$ & $\frac{1}{1024}\ \mathrm{cycles/FLOP}$ \\
$d_{\text{max}}$ & $20\ \mathrm{m}$                & $B_c,B_s$      & $500\ \mathrm{kHz}$ \\
$p_k^{\text{max}}$ & $41.76\ \mathrm{dBm}$          & $p_{\text{th}}^s,p_{\text{th}}^f$ & $46.99\ \mathrm{dBm}$ \\
\bottomrule
\end{tabular}
\end{table}

The performance of SflLLM is evaluated on a natural language generation task using the E2E dataset~\cite{novikova2017e2e}, which includes restaurant-domain data comprising approximately 42,000 training, 4,600 validation, and 4,600 test samples. The GPT-2 architecture~\cite{radford2019language} is employed, using its smallest variant, GPT2-S, featuring 12 Transformer decoder layers and approximately 124 million parameters. The training workload is estimated by assuming the backward pass requires double the computation of the forward pass; the embedding and positional encoding are neglected due to their minimal complexity.

Unless otherwise specified, we adopt a mini-batch size of 16, a learning rate of 0.0004, and a maximum sequence length of 512 for GPT2-S. For GPT2-M, the batch size is 12. LoRA modules are applied to the query and value matrices across all Transformer layers.

\begin{table}[!t]
\renewcommand\arraystretch{1.15}
\caption{Computational Complexity Analysis of GPT2-S with LoRA}
\label{tab_gpt2_final}
\centering
\footnotesize
\begin{tabular}{@{}lcc@{}}
\toprule
\textbf{Component} & \textbf{Parameters} & \textbf{FLOPs (GFLOP)} \\ 
\midrule
Token Embedding & 38.6M & -- \\
Position Encoding & 0.786M & -- \\
\hline
\textbf{Transformer Block $\times$12} & & \\ 
\quad LayerNorm & 1.5K & 0.025 \\ 
\quad Multi-Head Attention & 2.36M & 257.7 \\
\quad LoRA Adapter (per rank) & 1.5K & 0.050 \\ 
\quad Feed-Forward & 4.72M & 309.2 \\ 
\hline
Final LayerNorm & 1.5K & 0.025 \\ 
LM Head & -- & 1264.1 \\ 
\bottomrule
\end{tabular}
\end{table}

\subsection{Performance Evaluation of the Proposed SflLLM Framework}
To assess the effectiveness of SflLLM, we compare its performance against a centralized fine-tuning baseline, where raw data is aggregated at the server and processed using centralized LoRA training.

\subsubsection{Convergence Rate}
\begin{figure}[!t]
    \centering
    \subfloat[Validation loss with GPT2-S.]{%
        \includegraphics[width=0.48\textwidth]{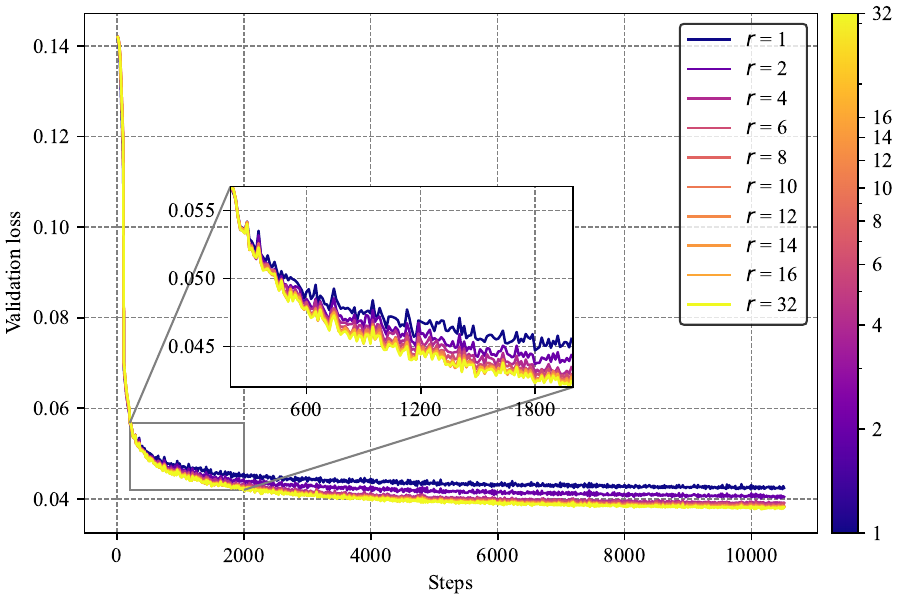}%
        \label{gpt2s}%
    }\hfill
    \subfloat[Validation loss with GPT2-M.]{%
        \includegraphics[width=0.48\textwidth]{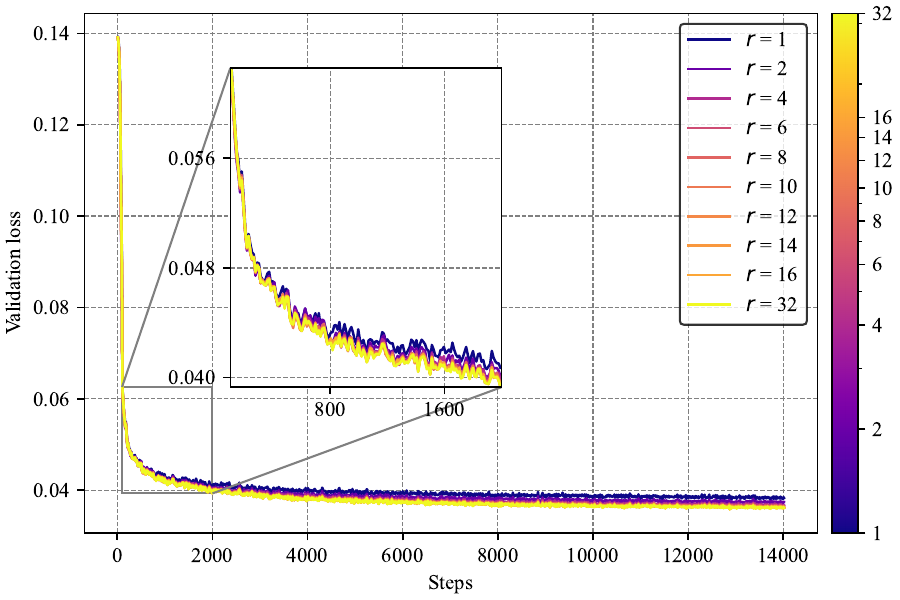}%
        \label{gpt2m}%
    }
    \caption{Validation loss on the E2E dataset with GPT2-S and GPT2-M for different LoRA ranks.}
    \label{gpt2}
\end{figure}

\begin{figure}[!t]
    \centering
    \includegraphics[width=3.5in]{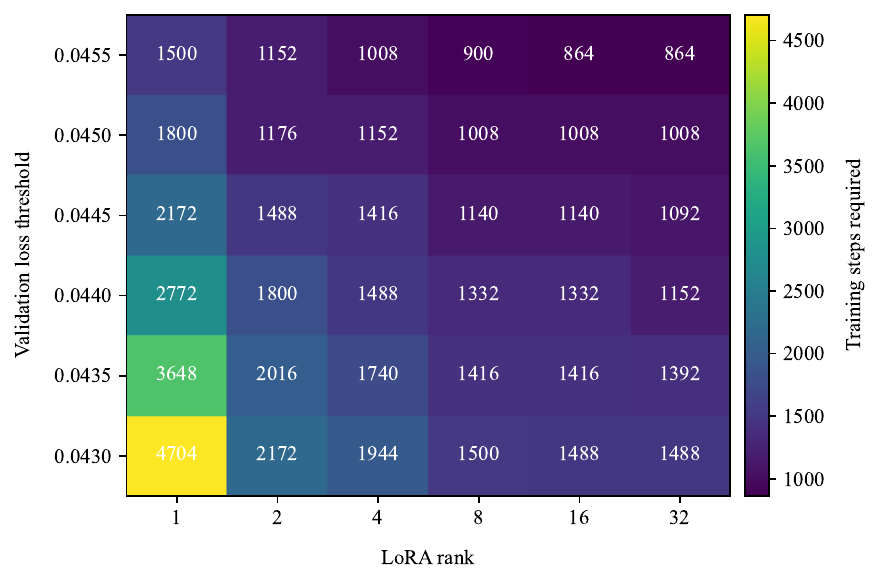}
    \caption{Steps required to achieve target validation loss under varying LoRA ranks.}
    \label{fig_heatmap}
\end{figure}

Fig.~\ref{gpt2} shows convergence trends for SflLLM on GPT2-S and GPT2-M. Validation performance is measured every 12 steps. Generally, higher LoRA ranks accelerate convergence, though benefits diminish beyond a certain point due to increased parameter count. Notably, Fig.~\ref{fig_heatmap} shows that fewer steps are needed to achieve a given loss with higher ranks. Rank selection directly impacts computational cost, communication overhead, and convergence rate, hence plays a vital role in minimizing training delay.

\subsubsection{Converged Accuracy}

Table~\ref{tab:rank_performance} compares final test Perplexity (PPL) values between SflLLM and centralized training for GPT2-S. SflLLM achieves comparable performance, with maximal PPL deviation within 0.001. This demonstrates SflLLM’s robustness to data heterogeneity, owing to its split model design—server-side models aggregate knowledge across clients, while client-side components are locally adapted. Higher ranks yield better PPL as they offer more trainable parameters, enhancing model expressiveness.

\begin{table}[t]
\scriptsize
\caption{Converged Test Perplexity for E2E}
\label{tab:rank_performance}
\centering
\setlength{\tabcolsep}{3pt}
\begin{tabular}{lccccc}
\toprule
\multirow{2}{*}{\textbf{Learning Framework}} & \multicolumn{5}{c}{\textbf{Rank}} \\
\cmidrule(lr){2-6}
 & 1 & 2 & 4 & 6 & 8 \\
\midrule
\textbf{Centralized} & 1.0424 & 1.0407 & 1.0393 & 1.0388 & 1.0385 \\
\textbf{SflLLM} & 1.0433 & 1.0413 & 1.0399 & 1.0398 & 1.0392 \\
\bottomrule
\end{tabular}
\end{table}

\subsection{Performance Evaluation of the Proposed Resource Management Strategy}
We evaluate the proposed resource allocation and rank selection strategy against four baselines:

\begin{itemize}
    \item \textbf{Baseline a}: Random subchannel allocation and PSD, random rank and split location.
    \item \textbf{Baseline b}: Random subchannel and PSD; proposed rank and split location selection.
    \item \textbf{Baseline c}: Random split location; proposed subchannel, power control, and rank selection.
    \item \textbf{Baseline d}: Proposed subchannel, power control, and split location; random rank selection.
\end{itemize}

\begin{figure}[!t]
    \centering
    \includegraphics[width=3.5in]{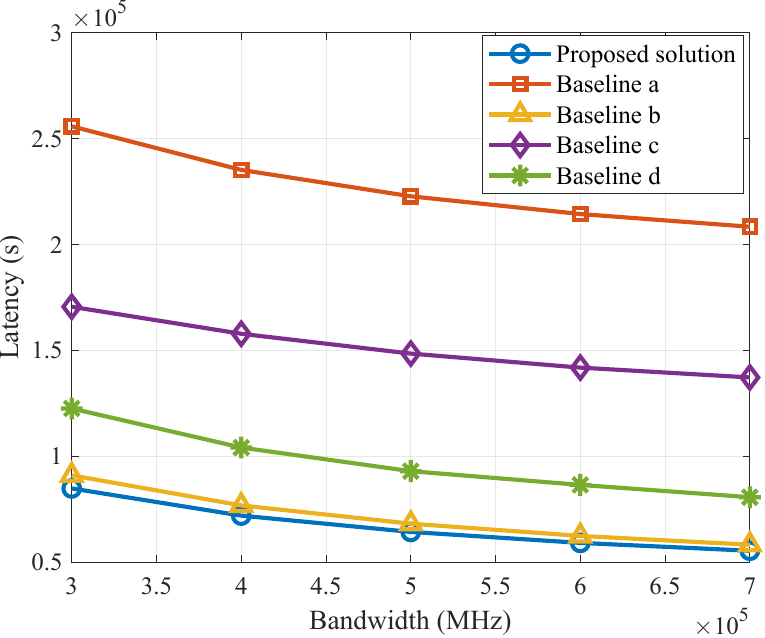}
    \caption{Total latency versus total bandwidth of each client.}
    \label{T_B}
\end{figure}

Fig.~\ref{T_B} illustrates the total training latency versus bandwidth. Our approach achieves up to 60\% latency reduction compared to baseline a. As bandwidth increases, communication delay decreases, shifting the bottleneck toward computation, hence reducing the performance gap with baseline b. The comparison with baseline d further emphasizes the importance of rank optimization.

\begin{figure}[!t]
    \centering
    \includegraphics[width=3.5in]{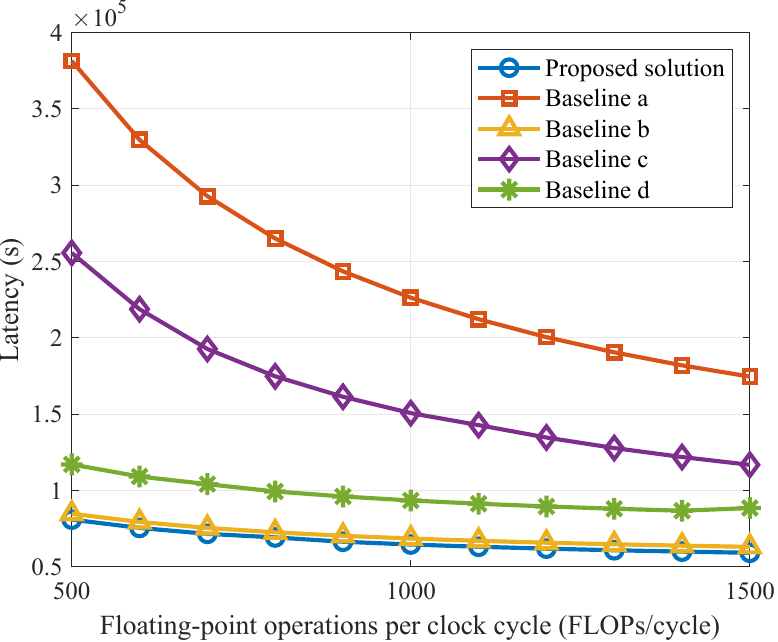}
    \caption{The effect of LoRA rank on the number of steps required to reach the target loss value.}
    \label{T_kappa}
\end{figure}

Fig.~\ref{T_kappa} shows latency as a function of the clients’ computational power (FLOPs per cycle). As expected, latency drops with higher compute power. The gap between our approach and baseline c narrows, highlighting that optimized split location becomes less impactful as computation dominates.

\begin{figure}[!t]
    \centering
    \includegraphics[width=3.5in]{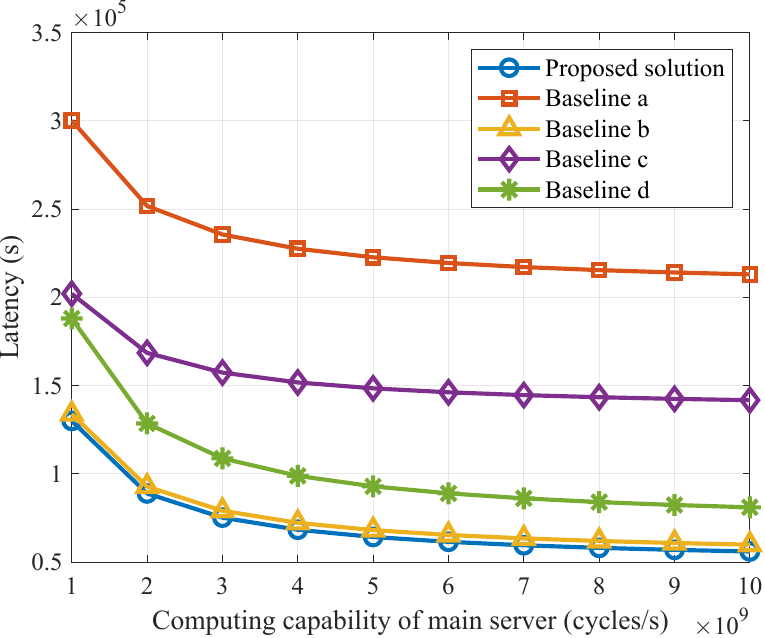}
    \caption{The effect of LoRA rank on the number of steps required to reach the target loss value.}
    \label{T_fs}
\end{figure}

Fig.~\ref{T_fs} presents latency trends with respect to main server computation power. Latency consistently improves with higher server capacity. Notably, the gap between baselines b and d indicates rank optimization contributes more to latency reduction than communication tuning in this setup.

\begin{figure}[!t]
    \centering
    \includegraphics[width=3.5in]{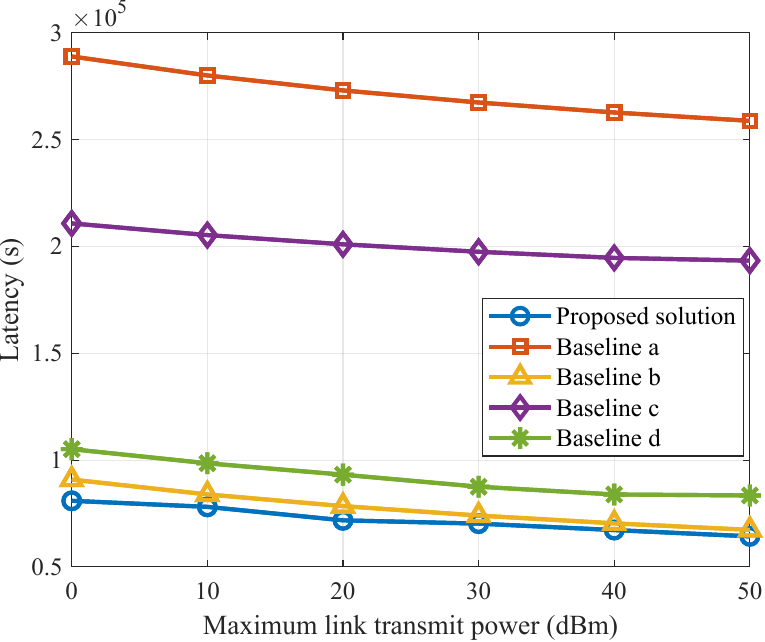}
    \caption{The effect of LoRA rank on the number of steps required to reach the target loss value.}
    \label{T_p}
\end{figure}

Finally, Fig.~\ref{T_p} shows that higher transmit power reduces delay across all schemes, with our proposed strategy offering the lowest latency. The benefit of power optimization becomes especially pronounced in bandwidth-limited scenarios.

\section{Conclusion}\label{Conclusion}

In this paper, we have proposed a novel split federated learning fine-tuned large language model framework (SflLLM) designed for distributed training of large models on resource-constrained edge devices. SflLLM reduces the client computational workload by splitting the model, introducing a federated server for parallel training, and aggregating client models. The computational and communication costs are further reduced by LoRA, significantly accelerating training by decreasing latency. Considering the heterogeneity of client communication conditions and computational capabilities, as well as the substantial impact of rank on convergence speed and training overhead, we proposed a joint resource allocation and rank selection scheme. This scheme jointly optimizes subchannel allocation, power control, split location, and rank selection to enhance the training speed of SflLLM over wireless networks. For the mixed-integer nonlinear programming delay minimization problem, we decomposed it into four sub-problems based on different decision variables and proposed solutions for each sub-problem. Subsequently, we presented a BCD-based algorithm to solve this optimization problem. Simulation results demonstrated that SflLLM achieves performance similar to centralized learning, and the proposed resource allocation and rank selection schemes significantly reduced training latency across different settings compared to traditional methods.

The findings of this paper highlight the potential of applying split federated learning to large model fine-tuning. Future research should focus on analyzing the effect of rank selection on convergence through theoretical derivations. Additionally, exploring an energy-efficient SflLLM framework and extending it to different types of pre-trained models are promising directions for further investigation.



\bibliographystyle{IEEEtran}
\bibliography{new_ref}


 




\vfill

\end{document}